\documentclass{article}
\usepackage{spconf,amsmath,graphicx}
\usepackage{amssymb}
\usepackage[dvipsnames]{xcolor}
\usepackage{hyperref}
\usepackage{tabularx}
\usepackage{multirow}

\def\x{{\mathbf x}}
\def\y{{\mathbf y}}
\def\h{{\mathbf h}}
\def\vi{{\mathbf{v}_\text{i}}}

\def\vih{{\widehat{\mathbf{v}_\text{i}}}}
\def\vo{{\mathbf{v}_\text{o}}}

\def\R{{\mathbb R}}
\def\tmax{{T_\text{max}}}
\def\umax{{U_\text{max}}}

\title{Extreme Encoder Output Frame Rate Reduction: Improving Computational Latencies of Large End-to-End Models}
\name{\begin{tabular}{c}Rohit~Prabhavalkar$^*$,~Zhong~Meng$^*$,~Weiran~Wang,~Adam~Stooke,~Xingyu~Cai\\~Yanzhang~He,~Arun~Narayanan,~Dongseong~Hwang,~Tara~N.~Sainath, Pedro~J.~Moreno\end{tabular}\thanks{$^*$Equal contribution. We would like to thank Shun Yao, Hong Jiao, Nanxin Chen, and Oren Litvin for helpful comments on this work.}}
\address{Google LLC, USA \\ 
\fontsize{9}{9}\selectfont\ttfamily\upshape {\texttt{\{prabhavalkar,zhongmeng\}@google.com}}}
\begin{document}
\ninept
\maketitle
\begin{abstract}
The accuracy of end-to-end (E2E) automatic speech recognition (ASR) models continues to improve as they are scaled to larger sizes, with some now reaching billions of parameters.
Widespread deployment and adoption of these models, however, requires computationally efficient strategies for decoding. 
In the present work, we study one such strategy: applying multiple frame reduction layers in the encoder to compress encoder outputs into a small number of output frames. 
While similar techniques have been investigated in previous work, we achieve dramatically more reduction than has previously been demonstrated through the use of multiple funnel reduction layers.
Through ablations, we study the impact of various architectural choices in the encoder to identify the most effective strategies.
We demonstrate that we can generate one encoder output frame for every 2.56 sec of input speech, without significantly affecting word error rate on a large-scale voice search task, while improving encoder and decoder latencies by 48\% and 92\% respectively, relative to a strong but computationally expensive baseline.
\end{abstract}
\begin{keywords}
end-to-end ASR, computational latency, runtime efficiency, large models
\end{keywords}
\section{Introduction}
\label{sec:intro}
End-to-end (E2E) approaches to automatic speech recognition (ASR)~\cite{li2022:e2ereview,prabhavalkar2023:e2ereview} have become increasingly widespread over the last few years; 
recurrent neural network transducers (RNN-Ts)~\cite{graves2013:rnnt, zhang2020:transformer_transducer}\footnote{We use the term RNN-T to refer to all models in this family, even those which employ non-recurrent encoders (e.g., a transformer~\cite{vaswani2017:transformer}).} 
and attention-based encoder-decoder models (AED)~\cite{chan2016:las} feature prominently in ASR research.
Recent years have witnessed a trend towards 
extremely large ASR models, in the range of a billion parameters or more~\cite{pratap2020:mmasr, li2021:mmasr_paper, radford2023:whisper, zhang2023:google_usm}.
Larger model sizes, coupled with full-sequence processing of the input audio, have enabled significant improvements in word error rate (WER), albeit at the cost of much higher computational latency.\footnote{We distinguish \emph{computational latency} -- the time required to process the end-pointed audio and produce a hypothesis -- from \emph{user-perceived latency}, which also includes additional delays in detecting the end of the utterance in order to close the microphone.}
While high-latency processing is acceptable for some tasks (e.g., offline video captioning), many tasks (e.g., recognizing short voice search queries) require extremely low user-perceived latency~\cite{shangguan2021:user_perceived_latency}.
Such tasks cannot benefit from large, full-sequence ASR models unless we develop new solutions to reduce computational latency.

We study the application of large E2E models ($\sim$900M parameter hybrid autoregressive transducer (HAT) models~\cite{variani2020:hat}, in this work) on a large-scale voice search task, and consider techniques to improve computational latency without sacrificing WER.
Our central insight is the following: the overall cost of decoding in HAT models is proportional to the maximum number of encoder output frames ($\tmax$) and the maximum number of non-blank outputs ($\umax$) produced by the model -- i.e., the cost scales as $O(\tmax + \umax)$ (details in Sec.~\ref{sec:decoding_cost}).
Thus, a large reduction in the number of encoder output frames, $\tmax$, enables large reductions in computational latency.
While the use of sub-sampling layers in the encoder is a standard strategy for E2E ASR~(e.g.,~\cite{chan2016:las, he2019:mobile_rnnt}), the novelty of our work lies in the scale of the reduction: we demonstrate that we can produce a single encoder output frame for every 2.56 sec of input speech, while only incurring a 3\% WER degradation relative to a strong, but computationally expensive and thus infeasible baseline, reducing computational latency by $82\%$. 
On our voice search task (cf. Table~\ref{tbl:testset-stats}), this reduction corresponds to decoding 3 frames or fewer for more than 95\% of the utterances -- a 64$\times$ reduction in the number of encoder output frames relative to the system in~\cite{gulati2020:conformer}.
The proposed techniques shift the model from a regime where $\tmax \gg \umax$ to one where $\tmax < \umax$, thus enabling large computational speed-ups. 

The contributions of our work are as follows: First, we demonstrate the effectiveness of \emph{extreme} reductions in the number of encoder output frames for E2E models through the use of multiple funnel reduction layers~\cite{dai2020:funnel_transformer}, which enable massive reductions in computational latency.
Since multiple encoder configurations can achieve the same amount of reduction, we investigate and determine the most effective configurations to balance latency and WER. 
Finally, in contrast to previous work on voice search tasks~\cite{ghodsi2020:stateless, prabhavalkar2021:less_is_more}, we demonstrate that for models with extreme encoder output reduction, increasing prediction network context improves accuracy and is critical to compensate for the information compression in the encoder.

\vspace{-0.1in}
\section{Related Work}
\label{sec:related_work}
The use of sub-sampling to reduce encoder output frames is a common strategy for E2E ASR: e.g., concatenating adjacent frames at intermediate encoder layers~\cite{he2019:mobile_rnnt}; convolutional sub-sampling layers~\cite{zhang2023:google_usm, Baevski2020:wav2vec2}; pyramidal bi-directional LSTM encoders~\cite{chan2016:las}; and using either a single~\cite{Ding2022:cascaded_encoder_system_paper}, or multiple~\cite{wang2023:massive_e2e_usm, cai2023:efficient_cascaded} funnel reduction layers~\cite{dai2020:funnel_transformer}.
While our work uses similar techniques, we explore encoder frame reductions that are an order of magnitude greater than in previous works.
We demonstrate that such extreme reduction is viable for low latency decoding in a voice search task. 
Techniques such as distillation~(e.g.,~\cite{kurata2020:knowledge_distillation}), quantization (e.g.~\cite{rybakov2023:2bit}), or (structured) parameter sparsity~(e.g.,~\cite{price2016:quantization_sparsity}) can all improve computational latency, and are equally applicable to the models proposed in this work.

\section{E2E ASR Model}
\label{sec:model}
The input speech, parameterized into features, is denoted as: $\x = (\x_1,  \ldots, \x_{T'})$, where $\x_t \in \R^{d}$. 
The corresponding transcript is denoted by $\y$, which is tokenized into a sequence of length $U$: $\y = (\y_1, \ldots, \y_{U})$. 
The E2E encoder transforms the input speech into a higher-level representation: $\h = (\h_1, \ldots, \h_T)$, where $\h_t \in R^{m}$. 
We refer to the ratio of the number of input frames to the number of output frames as the \emph{encoder reduction ratio}: $r_\text{enc} = \frac{T'}{T}$, and the effective amount of speech corresponding to each output frame as the \emph{encoder output duration}, $f_\text{enc}$.\footnote{For example, $f_\text{enc}=20$ms for an encoder which operates on 10ms input features, $\x_{t}$, with an encoder reduction ratio,  $r_\text{enc}=2$.}

\begin{figure}
  \centering
    \includegraphics[width=\columnwidth]{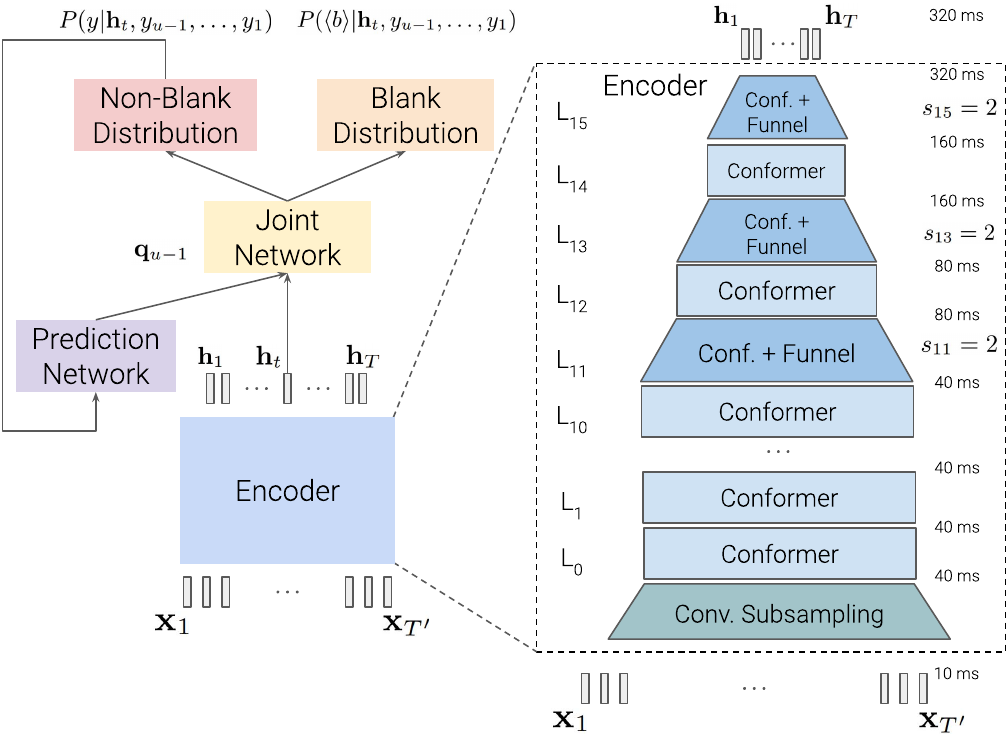}
    \vspace{-0.3in}
  \caption{The hybrid autoregressive transducer (HAT)~\cite{variani2020:hat} (left).
  We replace some of the layers with the funnel reduction variant described in Eq.~\eqref{eq:funnel}. The encoder structure (right) -- ($s^2_{15}, s^2_{13}, s^2_{11}$) (see notation in Sec.~\ref{sec:experimental_setup}) -- corresponds to an encoder reduction factor, $r_\text{enc}=8$, with an encoder output duration, $f_\text{enc}=320$ms.}
  \label{fig:hat}
  \line(1,0){\columnwidth}
  \vspace{-0.25in}
\end{figure}
We conduct experiments using the hybrid autoregressive transducer (HAT)~\cite{variani2020:hat} as a representative E2E model.
Of the two other prominent E2E model types, AED can also accomodate extreme frame reduction; connectionist temporal classification (CTC)~\cite{graves2006:ctc} cannot, since it outputs only one non-blank token per frame and thus requires $T \geq U$.
The HAT model, illustrated in Fig.~\ref{fig:hat}, consists of an encoder; a prediction network; a joint network; and two output distributions over blank and non-blank symbols, respectively.
Details of HAT models are omitted due to space constraints; interested readers are directed to~\cite{li2022:e2ereview, prabhavalkar2023:e2ereview,variani2020:hat}.

\vspace{-0.1in}
\section{Proposed Methods}
We focus on the two aspects 
most relevant to this work: techniques for achieving large encoder reduction ratios; and 
decoding strategies which can be implemented efficiently on hardware accelerators such as Tensor Processing Units (TPUs)~\cite{jouppi2021:tpu}. 

\vspace{-0.1in}
\subsection{Encoder Structure}
Our encoder corresponds to the conformer architecture proposed by Gulati et al.~\cite{gulati2020:conformer}, illustrated in Fig.~\ref{fig:hat}.
The encoder consists of a convolutional sub-sampling block which increases the output duration to 40ms, 
followed by a series of 16 conformer blocks,
with one modification
: following~\cite{Li21Better} we interchange the order of the convolution and multi-headed self-attention (MHSA).
Denoting the conformer block inputs by $\vi$, the outputs, $\vo$, are computed as: 
\begin{align}
    \vi' &= \vi + \frac{1}{2} \text{FFN}_1(\vi) + \text{Conv}(\vi + \frac{1}{2} \text{FFN}_1(\vi)) \nonumber \\
    \vi''  &= \vi' + \text{MHSA}(\text{Q}=\vi', \text{KV}=\vi') \label{eq:conformer_mhsa} \\
    \vo   &= \text{LayerNorm}(\vi'' + \frac{1}{2} \text{FFN}_2(\vi'')) \label{eq:conformer}
\end{align}
where, $\text{FFN}(\cdot)$ represents a feed-forward network; $\text{MHSA}(\text{Q}=\x, \text{KV}=\y)$ represents multi-headed self-attention~\cite{vaswani2017:transformer}; 
$\text{Conv}(\cdot)$ refers to the convolution module from~\cite{gulati2020:conformer}; and $\text{LayerNorm}(\cdot)$ refers to layer normalization. 
Thus in the vanilla conformer, the number of frames in the conformer output, $|\vo|$, is exactly the same as the number of input frames, $|\vi|$.

In this work, we progressively reduce the number of frames in the conformer blocks using funnel reduction proposed by Dai et al.~\cite{dai2020:funnel_transformer} which uses pooling for the query in MHSA:
\begin{subequations}
\begin{align}
    \vih &= \text{AvgPooling}(\vi', \text{query-stride}=s) \\
    \vih' &= \text{MaxPooling}(\vi', \text{query-stride}=s) \\
    \vi'' &= \vih' + \text{MHSA}(\text{Q}=\vih, \text{KV}=\vi') 
\end{align}
\label{eq:funnel}
\end{subequations}
\noindent where, $\text{\{Avg/Max\}Pooling}(\vi, \text{query-stride}=s)$ corresponds to average or max pooling, over non-overlapping blocks of length $s$ (query stride).
Pooling reduces the effective length of $\vih$ and $\vih'$ (and thus the output) by a factor, $s$: $|\vo| = \frac{1}{s} |\vi|$. 
The use of funnel reduction layers does not change the number of model parameters.

\textbf{Obtaining Large Encoder Reduction Factors: }
In our work, we replace the standard conformer blocks, with the funnel reduction-based variant described in Eq.~\eqref{eq:funnel}, as illustrated in Fig.~\ref{fig:hat}.
Assuming that this modification is applied to M conformer blocks, with individual query strides $s_i$, the total encoder reduction ratio is: $r_\text{enc} = \prod_{i=1}^M s_i$. 
Different configurations of funnel reduction layers and query strides can result in the same encoder reduction factor.
Models with the same reduction factor will have the same decoder latency, since both process the same number of encoder output frames; adding funnel reduction layers earlier, and/or using larger query strides will result in lower encoder latency by reducing the amount of computation in the higher conformer layers.

\begin{table}
  \centering
  \begin{tabular}{|c|c|c|c|}
    \hline
    Set & Utts & Utt. Len. (sec) & Token Len. \\
        &            & P50 / P90 / P95 / P99 & P50 / P90 / P95 / P99 \\
    \hline
    VS &  8884  & 4.0 / 6.5 / 7.6 / 10.0 & 8 / 15 / 17 / 23 \\
    RM & 10000  & 5.7 / 6.9 / 7.4 / 8.2  & 8 / 13 / 15 / 20 \\ 
    RQ & 10000  & 5.4 / 6.0 / 6.2 / 6.7  & 7 / 10 / 11 / 14 \\ 
    \hline
  \end{tabular}
  \caption{Percentile statistics for test sets that we report results on.}
  \label{tbl:testset-stats}
  \line(1,0){\columnwidth}
  \vspace{-0.3in}
\end{table}
\vspace{-0.1in}
\subsection{Decoder Latency Improvements via Encoder Reduction}
\label{sec:decoding_cost}
Broadly speaking, RNN-T/HAT models can be decoded using two possible strategies: frame-synchronous~\cite{graves2013:rnnt}, or alignment-length synchronous~\cite{saon2020:label_synch_rnnt} search. 
While decoder latency improvements can be obtained through either strategy with the proposed encoder reduction, the alignment-length synchronous strategy has advantages for TPU-based deployment, and this is the strategy adopted in this work.

The original RNN-T decoding algorithm proposed by Graves~\cite{graves2013:rnnt} can be termed frame-synchronous: all hypotheses in the beam correspond to the same encoder output frame; different hypotheses in the beam might contain a different number of non-blank labels.
Thus, a ``step" (i.e., processing one encoder output frame completely, before moving to the next frame) 
has variable computational latency, since each beam produces a variable number of outputs at each step. 

An alignment-length synchronous strategy~\cite{saon2020:label_synch_rnnt}, however, ensures that all hypotheses in the beam contain the same number of labels (blank + non-blank).
A single ``step" in such a strategy is far simpler, and more amenable to implementations on hardware accelerators: for a beam of size K, we extend every hypothesis in the beam with a single label (blank or non-blank); we then retain the top-K lowest cost extended hypotheses, which form the beam for the next step. 
Decoding terminates when the lowest cost hypothesis contains a blank label produced from the last frame.
Crucially, such a search can be trivially batched across the individual beams of multiple input utterances, thus improving system throughput.
In summary, each such step in an alignment-length synchronous search has a fixed computational cost, denoted by $C_\text{exp}$. 

Assuming that the longest input speech utterance has length $T'_\text{max}$, the longest encoder output sequence is $T_\text{max} = \frac{T'_\text{max}}{r_\text{enc}}$.
We can further choose an upper bound, $U_\text{max}$, corresponding to the largest number of non-blank symbols for the evaluation task. 
With these settings, the total computational cost of the decoding is  $C_\text{exp} (T_\text{max} + U_\text{max}) = C_\text{exp} (\frac{T'_\text{max}}{r_\text{enc}} + U_\text{max})$, which is a tight bound.
Since $\frac{T'_\text{max}}{r_\text{enc}} \ll U_\text{max}$ with extreme encoder reduction, the actual savings in decoder computation can be very significant, as demonstrated by our experimental results (cf. Table~\ref{tbl:results2}).

\vspace{-0.1in}
\section{Experimental Setup}
\label{sec:experimental_setup}
We evaluate the effectiveness of the proposed methods on a large-scale voice search task, as detailed below.

\textbf{Training and Evaluation Sets:}
The voice search training set consists of 520M utterances, extracted from Google voice search traffic, totalling 490K hours of speech, with an average duration of 3.4 sec.
A small percentage of utterances are anonymized and human-transcribed; the majority of utterances are pseudo-labeled using a teacher model~\cite{hwang2022:pseudolabel}.
All data processing abides by Google AI principles~\cite{google_ai_principles}.

We report results on multiple test sets, to measure the impact of the proposed techniques.
The voice search test set (VS), corresponds to the ``head" of the utterance distribution.
We also evaluate performance on test sets which contain rare words from specific domains --  
maps (RM), and search queries (RQ) 
-- generated using a text-to-speech (TTS) system as described in~\cite{peyser2020:rare_word_mwer}.
These sets capture performance on the ``tail" of the utterance distribution.
Detailed statistics 
can be found in Table~\ref{tbl:testset-stats}.
Although voice search queries are generally quite short (only 5\% of the queries are longer than 7.6 sec), many utterances contain a fairly large number of output tokens.

\textbf{Input Processing and Data Augmentation: }
The input speech is parameterized into 128 dimensional log mel filterbank features (window size: 32ms; stride: 10ms). 
To improve robustness, we use various data augmentation techniques: artificially simulating noisy utterances~\cite{kim2017:mtr}; mixed-bandwidth training with equal probability downsampling to 8KHz or 16KHz~\cite{li2012:mixed_bandwidth}; and SpecAug~\cite{park2019:specaug}. 

\textbf{Encoder Architecture:} All models in this work are trained with the encoder configuration described in Fig.~\ref{fig:hat}.
It uses the convolutional sub-sampling block which increases the frame duration to 40ms.
This is followed by a sequence of 16 conformer blocks, with 1536 units in each block; multi-headed self-attention with 8 attention heads; and a convolutional kernel size of 15.
This forms our baseline configuration (B0).

We establish a shorthand notation -- e.g., ($s_{15}^4, s_{0}^2$) -- to represents a system with funnel reduction layers added at the last ($\text{L}_{15}$) and first ($\text{L}_0$) layers in the conformer stack, with query strides 4, and 2, respectively.

\textbf{Decoder Architecture:} The joint network uses the standard $\tanh(\cdot)$ combination~\cite{graves2013:rnnt} after linearly projecting the encoder and prediction network outputs to 640 dimensions.
All models tokenize the transcripts using 4096 word-pieces~\cite{Schuster2002:wordpieces}. 
Most of our models use a $V^{2}$ 
embedding prediction network~\cite{rami21:tied_reduced_rnnt}, which concatenates and projects the embeddings of the last two predicted non-blank labels. 
Since we compress information along the time-axis in the encoder, we re-evaluate the importance of label conditioning for models with extreme encoder frame rate reduction by also evaluating an LSTM-based~\cite{HochreiterSchmidhuber97:LSTM} prediction network: a two layer network with 2048 cells per layer (LSTM 2x2048).
Using an LSTM-based decoder increases the number of parameters by 20M over the system with a $V^2$ embedding network.

\textbf{Training and Inference: } Models are trained with an AdaFactor optimizer~\cite{shazeer2018:adafactor}, and a transformer learning rate schedule~\cite{vaswani2017:transformer}, over mini-batches of 4096 utterances. 
All models are built using the Lingvo toolkit~\cite{shen2019:lingvo}, and are trained and decoded on TPUv3 chips~\cite{jouppi2021:tpu}.
We report computational latencies by benchmarking the models with a batch size of 8 utterances, assuming a maximum audio input length of $T'_\text{max}=15.36$ sec, on a single TPUv3 chip~\cite{jouppi2021:tpu}, assuming $U_\text{max}=30$.\footnote{Since the actual mix of utterance lengths in the batch at runtime is unknown, we report decoder latency by estimating $C_\text{exp}$, and then computing $C_\text{exp}(\umax + \tmax)$. Encoder latency is computed as the time required to encode a batch of utterances of length $T'_\text{max}$.}

\vspace{-0.2in}
\section{Results}
\label{sec:results}
\begin{table}
  \centering
  \begin{tabular}{|c|r|c|c|c|c|}
    \hline
    ID & Encoder Config & $f_\text{enc}$ & \multicolumn{3}{c|}{WER (\%)} \\
    \cline{4-6}
       & & (ms)    & VS & RM & RQ  \\
    \hline
    B0 &  -  & 40 & \textbf{3.6} & \textbf{11.9} & \textbf{19.9} \\
    \hline
    E1 &  $s_{15}^2$  & 80 & 3.7 & 12.3 & 20.1 \\
    \hline
    E2 & $s_{13}^2, s_{15}^2$  & 160 & \textbf{3.6} & 12.5 & 20.3 \\
    \hline
    E3 & $s_{11}^2, s_{13}^2, s_{15}^2$  & 320 & 3.7 & 12.6  & 20.4\\
    \hline
    E4 &  $s_9^2, s_{11}^2, s_{13}^2, s_{15}^2$  & 640 & 3.9 & 13.0 & 21.2\\
    \hline
    E5 &  $s_7^2, s_9^2, s_{11}^2, s_{13}^2, s_{15}^2$  & 1280 & 3.7 & 12.7 & 21.0 \\
    \hline
    E6 & $s_5^2, s_7^2, s_9^2, s_{11}^2, s_{13}^2, s_{15}^2$  & 2560 & 4.0 & 13.5 & 21.0 \\
    \hline
    E7 & $s_3^2, s_5^2, s_7^2, s_9^2, s_{11}^2, s_{13}^2, s_{15}^2$  & 5120 & 5.4  & 15.4 & 22.8 \\
    \hline
  \end{tabular}
  \caption{Impact of using multiple funnel transformer layers in the conformer architecture.}
  \label{tbl:results1}
\end{table}
\begin{table}
  \centering
  \begin{tabular}{|c|r|c|c|c|c|}
    \hline
    ID & Encoder Config & $f_\text{enc}$ & \multicolumn{3}{c|}{Comp. Latency (ms)} \\
    \cline{4-6} 
    & & (ms) & Enc & Dec & Total  \\
    \hline
    B0 &  -  & 40 & 144 & 526 & 670  \\
    \hline
    E1 &  $s_{15}^2$  & 80 & 142 & 280 & 423 \\
    \hline
    E2 &  $s_{13}^2, s_{15}^2$  & 160 & 134 & 155 & 290 \\
    \hline
    E3 &  $s_{11}^2, s_{13}^2, s_{15}^2$  & 320 & 121 & 96 & 217 \\
    \hline
    E4 & $s_9^2, s_{11}^2, s_{13}^2, s_{15}^2$  & 640 & 106 & 66 & 172 \\
    \hline
    E5 &  $s_7^2, s_9^2, s_{11}^2, s_{13}^2, s_{15}^2$  & 1280 & 90 & 51 & 141 \\
    \hline
    E6 &  $s_5^2, s_7^2, s_9^2, s_{11}^2, s_{13}^2, s_{15}^2$  & 2560 & 74 & 44 & 118 \\
    \hline
    E7 &  $s_3^2, s_5^2, s_7^2, s_9^2, s_{11}^2, s_{13}^2, s_{15}^2$  & 5120 & \textbf{58} & \textbf{34} & \textbf{92} \\
    \hline
  \end{tabular}
  \caption{Computational latency benchmarked on a single TPUv3 chip~\cite{jouppi2021:tpu} for a batch of 8 utterances ($T'_\text{max}=15.36$ sec; $\umax=30$).} 
  \line(1,0){\columnwidth}
  \vspace{-0.4in}
  \label{tbl:results2}
\end{table}

\textbf{WER and Latency vs. Encoder Output Duration:}
We study the impact of adding funnel reduction to multiple conformer layers on WER and computational latency.
All models in the first set of experiments use $V^2$ embedding prediction networks.
We modify the encoder starting with the final conformer layer ($L_{15}$), progressively modifying every other conformer layer (i.e., $L_{13}, L_{11} \ldots$, etc.) with a query stride of 2 to gradually increase the encoder reduction rate.
All models have exactly the same number of parameters (880M).
As can be seen in Table~\ref{tbl:results1}, the WER on the VS test set is fairly stable across all systems, even with encoder frame durations as high as 1.28 sec; increasing reduction further degrades WER significantly (cf., E6, E7). 
Trends are similar across all rare word test sets: e.g., on RM we observe relative WER degradations ranging from 3.4\% -- 6.7\% as we increase encoder frame durations from E1 to E5.
While B0 achieves the best results in terms of WER, the computational latency metrics in Table~\ref{tbl:results2} present an entirely different story: a computational latency of 670ms for a response to a short query is unacceptable from a user-experience perspective; 
systems with higher encoder reduction ratios such as E5, provide a balance between WER and latency (79\% latency improvement over B0).
We study strategies to further mitigate the WER degradation for extreme frame rate reduction models, E5 and E6, in subsequent experiments.

\begin{table}
  \centering
  \begin{tabularx}{\columnwidth}{|p{0.5cm}|>{\centering\arraybackslash}X|p{3.1cm}|>{\centering\arraybackslash}X|>{\centering\arraybackslash}X|>{\centering\arraybackslash}X|>{\centering\arraybackslash}X|}
    \hline
    $f_\text{enc} $ & ID & Encoder & \multicolumn{3}{c|}{WER (\%)} & Enc. \\
                                         \cline{4-6}
        (ms)                &    & config &       VS & RM & RQ & (ms)                       \\
    \hline
    \multirow{5}{*}{1280} 
      & E5 & $s_7^2, s_9^2, s_{11}^2, s_{13}^2, s_{15}^2$ & \textbf{3.7} & \textbf{12.7} & \textbf{21.0} & 90 \\
      \cline{2-7}
      & E51 & $s_{11}^2, s_{12}^2, s_{13}^2, s_{14}^2, s_{15}^2$ & 3.9 & 13.4 & 21.9 & 116 \\
      \cline{2-7}
      & E52 & $s_{4}^2, s_{5}^2, s_{6}^2, s_{7}^2, s_{8}^2$ & 3.9 & \textbf{12.7} & \textbf{21.0} & \textbf{60} \\
      \cline{2-7}
      & E53 & $s_{14}^8, s_{15}^4$ & 3.8 & \textbf{12.7} & \textbf{21.0} & 133 \\
      \cline{2-7}
      & E54 & $s_{13}^4, s_{15}^8$ & 3.8 & 13.1 & 21.5 & 128 \\
    \hline
    \hline
    \multirow{5}{*}{2560}
     & E6 & $s_5^2, s_7^2, s_9^2, s_{11}^2, s_{13}^2, s_{15}^2$ & 4.0 & 13.5 & \textbf{21.0} & 74 \\
     \cline{2-7}
     & E61 & $s_{10}^2, s_{11}^2, s_{12}^2, s_{13}^2, s_{14}^2, s_{15}^2$ & 3.9 & 13.3 & 21.2 & 108 \\
     \cline{2-7}
     & E62 & $s_{4}^2, s_{5}^2, s_{6}^2, s_{7}^2, s_{8}^2, s_{9}^2$ & 4.3 & 14.0 & 21.9 & \textbf{59} \\
     \cline{2-7}
     & E63 & $s_{14}^8, s_{15}^8$ & \textbf{3.8} & 13.3 & \textbf{21.0} & 133 \\
     \cline{2-7}
     & E64 & $s_{13}^8, s_{15}^8$ & \textbf{3.8} & \textbf{13.1} & 21.2  & 126  \\
     \hline
  \end{tabularx}
  \caption{Choice of query stride and funnel transformer layer indices, for a fixed $f_\text{enc}$. Decoder latency is fixed for a given $f_\text{enc}$.}
  \line(1,0){\columnwidth}
  \vspace{-0.4in}
  \label{tbl:results3}
\end{table}
\textbf{Choice of Query Stride and Funnel Layer Indices:}
We perform ablations to determine the impact of where the funnel reduction layers are added, and the degree of reduction through the query stride to achieve a given amount of reduction.
Since the decoder latency is the same for models with the same encoder frame duration, $f_\text{enc}$, we only report encoder latency in the results.
We focus on the case of $f_\text{enc} = 1280$ms, and $2560$ms, since these offer the best computational latency with reasonable WER.
Results are presented in Table~\ref{tbl:results3}, where we observe a few trends. 
First, configurations which use more aggressive frame rate reduction for latter conformer layers can be just as effective as those that gradually reduce the frame rate over multiple layers (or slightly better); this is particularly true for models with larger $f_\text{enc}$ (cf., E64 vs. E6, or E61). However, these configurations have unacceptably high encoder latencies. 
For models with larger $f_\text{enc}$, adding funnel reduction layers towards the end of the conformer stack appears to perform better than adding these early in the network (e.g., E6, or E61 vs. E62).
Finally, two aspects where the trends differ based on the amount of reduction: 
adding reduction layers at the end, or in the middle of the network performs similarly for $f_\text{enc}=1280$ms (cf. E5 vs. E52), but reduction at the end appears to be preferable for high reduction rates (cf. $f_\text{enc}=2560$ms: E62 vs. E6); reductions at every other layer seem to be more effective than reductions at consecutive layers for $f_\text{enc}=1280$ms (cf. E5 vs. E51), but both choices perform similarly for $f_\text{enc}=2560$ms (cf. E6 vs. E61).
We use configurations E5 and E6 in subsequent ablations since they offer the best tradeoff between latency and WER (E52 degrades WER on the VS set by 5\%; E64 increases encoder latency by 70\%).

\begin{table}
  \centering
  \begin{tabular}{|c|c|c|c|c|c|c|}
    \hline
    $f_\text{enc} $ & ID & Pred. Network & \multicolumn{3}{c|}{WER (\%)} & Dec. \\
                                         \cline{4-6}
        (ms)                &    & config &       VS & RM & RQ & (ms)                       \\
    \hline
    \multirow{2}{*}{1280} 
      & E5 & $V^2$ & \textbf{3.7} & 12.7 & 21.0 & \textbf{51} \\
    \cline{2-7}
      & E5D1 & LSTM-2x2048 & 3.8 & \textbf{12.5} & \textbf{20.1} & 63\\
    \hline
    \hline
    \multirow{2}{*}{2560}
     & E6 & $V^2$ & 4.0 & 13.5 & 21.0 & \textbf{44} \\
     \cline{2-7}
     & E6D1 & LSTM 2x2048 & \textbf{3.7} & \textbf{12.5}  & \textbf{20.3} & 54 \\
     \hline
     \hline
     \multirow{2}{*}{5120}
     & E7 & $V^2$ & 5.4 & 15.4  & 22.8 & \textbf{34} \\
     \cline{2-7}
     & E7D1 & LSTM 2x2048 & \textbf{3.8} & \textbf{12.8}  & \textbf{20.7} & 42 \\
     \hline
  \end{tabular}
  \caption{Impact of prediction network on WER and decoder latency.}
  \vspace{-0.1in}
  \label{tbl:results4}
\end{table}
\vspace{0.2in}
\textbf{LSTM Prediction Networks Improve Extreme Reduction:} Previous works have demonstrated that using limited label context in the prediction network is sufficient for voice-search tasks~\cite{variani2020:hat, ghodsi2020:stateless, prabhavalkar2021:less_is_more, rami21:tied_reduced_rnnt}. 
These works truncate the label history when modeling the non-blank distribution to the last $N$ labels: $P(y | h_t, y_{u-1}, \ldots, y_{1}) \triangleq P(y | h_t, y_{u-1}, \ldots, y_{u-N})$. 
However, these previous works all study models with low encoder frame durations. 
We investigate the impact of prediction network context in Table~\ref{tbl:results4}.
For all models, the use of an LSTM network, which models additional context, improves performance over $V^2$ embedding networks, especially on the rare word test sets, with a $\sim$10ms increase in latency.
The trend is clearest for E7 -- the model with the highest encoder frame duration -- where the LSTM-based system improves WER relatively by 29.6\% on VS; 16.9\% on RM; and 9.2\% on RQ.
We interpret this result as follows: in models with a large number of output encoder frames, the model can rely more on the encoder and less on the prediction network; when the number of output encoder frames reduces, the ability to model richer context in the prediction network becomes crucial. 

\begin{table}
  \centering
  \begin{tabular}{|c|c|c|c|c|c|c|}
    \hline
    $f_\text{enc}$ & ID & \multicolumn{3}{c|}{WER (\%)} & Total Comp. \\
                                         \cline{3-5}
     (ms)  &  & VS & RM & RQ &  Latency (ms) \\
    \hline
    \multirow{2}{*}{40} & B0        &  \textbf{3.6} & 11.9 & 19.9 & \multirow{2}{*}{670} \\
    \cline{2-5}
                        & B0 + MWER & \textbf{3.6} & \textbf{11.7} & \textbf{19.4} &   \\
    \hline
    \hline
    \multirow{2}{*}{2560} & E6D1        &   \textbf{3.7} & 12.5 & 20.3 & \multirow{2}{*}{118} \\ 
    \cline{2-5}
                          & E6D1 + MWER & \textbf{3.7} & \textbf{12.1} & \textbf{19.6} &  \\
    \hline
  \end{tabular}
  \caption{Improvements with MWER~\cite{prabhavalkar2018:mwer} training.}
  \label{tbl:results5}
  \line(1,0){\columnwidth}
  \vspace{-0.25in}
\end{table}
\textbf{Improvements with MWER Training: } In our final experiments, we report the results of minimum word error rate (MWER) training~\cite{prabhavalkar2018:mwer} applied to the best performing system with $f_\text{enc}=2560$ms (E6D1), and compare it to the baseline (B0).
The MWER loss is computed using 4-best hypotheses, and is interpolated with the HAT loss (HAT loss scale: $0.03$) during training.
As can be seen in Table~\ref{tbl:results5}, MWER improves both models, but only for the rare word sets; gains are larger for the model with extreme encoder frame reduction.
Comparing the two systems after MWER training, the E6D1 system is only 2.5\% worse on VS, and 3.3\% worse on the rare word sets, relative to the baseline after MWER, but improves total computational latency by $82\%$ -- a user-perceived latency improvement of 550ms. 

\vspace{-0.2in}
\section{Conclusions}
\label{sec:conclusions}
In this work, we explored extreme encoder frame rate reduction in order to improve computational latencies of large E2E models.
Through experimental results, we demonstrate that we can decode speech by producing encoder outputs with an encoder duration of 2.56 sec through the use of multiple funnel reduction layers~\cite{dai2020:funnel_transformer} which progressively increase the encoder reduction rate.
The proposed techniques are simple to implement, providing new avenues for deploying large E2E ASR models with low computational latencies for voice search tasks.

\vfill\pagebreak

\bibliographystyle{IEEEbib}
\bibliography{Template}

\end{document}